\documentclass[letterpaper]{article} 
\usepackage[preprint]{aaai2027}  
\usepackage[hyphens]{url}  
\usepackage{graphicx} 
\usepackage{natbib}  
\usepackage{caption} 
\usepackage{booktabs}
\usepackage{array}
\usepackage{amsmath}
\usepackage{amssymb}
\usepackage{multirow}
\usepackage{pdfpages}

\title{Diagnosing Sampled LLM Reasoning in Formal Geometry: Coverage, Realization, and Validity Evidence}
\author{Xiao Yue\textsuperscript{\rm 1}, Guangzhi Qu\textsuperscript{\rm 1}}
\affiliations{\textsuperscript{\rm 1}Oakland University, Rochester, United States\\
\{xiaoyue, gqu\}@oakland.edu}

\begin{document}

\maketitle

\begin{abstract}
Repeated sampling can reveal a correct numerical answer without yielding either a reliable system output or a supported derivation. We present \emph{Coverage, Realization, and Validity Evidence} (CRV), an evaluation protocol for sampled large language model (LLM) reasoning over formal geometry states. Coverage is answer availability, realization is readout accuracy on the frozen candidate pool, and validity evidence is a label-blinded critic judgment of derivational support rather than a proof certificate. CRV freezes each candidate pool before comparing readouts and analyzes covered failures by correct-answer multiplicity and within-problem discrimination. On HardShift441, a 441-problem set for which a reference solver leaves 406 problems unsolved, a LoRA-adapted Qwen2.5-7B generator obtains 24.2\% average single-sample accuracy and 68.9\% pass@16, whereas verifier-weighted self-consistency (WSC) reaches 38.0\%. Readout accuracy is particularly low when the correct answer occurs only once or twice in the pool. In a separate constructed audit of 195 covered problems, the critic labels 12 correct-answer representatives as supported, 181 as refuted, and two as uncertain. These results show that coverage, realization, and validity evidence from the critic are distinct quantities and should be reported separately.
\end{abstract}

\section{Introduction}

LLM-based geometry solvers are often evaluated using a single generated numerical answer.  Answer-matching accuracy records whether this answer is correct, but not whether repeated sampling would produce the correct answer or whether the reasoning is supported by the given premises.  This limitation is especially important in geometry, where each step may depend on earlier deductions or on relations shown only in a diagram.

We study this issue in a common input--output setting designed to make the model's reasoning inspectable.  The model receives a problem together with a natural-language description of its \emph{initial formal state}, which contains the premises and constructions known before any solution steps are taken.  The model then produces a textual derivation followed by a numerical answer.  The state captures relations given in the problem text or diagram but excludes intermediate deductions from the reference proof.  Our main study focuses on reasoning generated in this setting before a symbolic solver or checker can correct model errors.

Even in this inspectable setting, accuracy computed from a single generated output conflates three questions.  First, does the generator produce the correct answer within a pool of sampled candidates?  We call a problem \emph{answer-level covered} when at least one candidate reaches the correct numerical answer.  Second, if a correct answer is present, does a majority-vote or verifier-based readout select it from the candidate pool \cite{cobbe2021training,wang2022self,brown2024large,snell2025scaling}?  Third, does a blinded audit find that a correct-answer trace is supported by the given premises?  We study coverage, realization, and validity evidence separately under a solver-unsolved hard-state shift.

To diagnose these failure modes, we introduce the \emph{Coverage, Realization, and Validity Evidence} (CRV) analysis framework.  CRV provides a controlled protocol for measuring the three components separately.  It freezes generated answer pools so that readouts operate on identical answer clusters, measures how much of the available coverage each readout realizes, and localizes unrealized coverage by correct-answer multiplicity and hardest-negative discrimination.  A separate blinded audit supplies validity evidence for constructed representatives from correct-answer and hard incorrect-answer clusters without revealing answer correctness to the critic.

We evaluate CRV on HardShift441, a 441-problem solver-unsolved hard-state shift derived from FormalGeo7K \cite{zhang2023formalgeo}. A reference FGeo-HyperGNet (hereafter HyperGNet) run \cite{zhang2024fgeo} leaves 406 of these problems unsolved. The set is disjoint from the legacy-hard evaluation set by problem identifier, but we do not claim that every problem is globally unseen. With a LoRA-adapted Qwen2.5-7B generator, average single-sample accuracy is 24.2\%, pass@16 is 68.9\%, and verifier-weighted self-consistency (WSC) reaches 38.0\%; readout accuracy is particularly low when the correct answer appears once or twice. In the constructed audit pool, the critic labels only 12 of 195 correct-answer representatives as supported. A second model agrees with the critic on 33 of 34 reviewed verdicts, and human review agrees with the critic on all 20 reviewed verdicts in the sampled strata; neither check estimates full-pool critic accuracy.

In summary, this paper makes the following main contributions:
\begin{itemize}
    \item We present CRV as a controlled evaluation protocol---not a new selector---that measures coverage, realization, and validity evidence from a blinded critic on frozen candidate pools.
    \item We localize the generation-to-selection gap on hard formal-geometry states to low-multiplicity pools and separate pointwise ranking from answer-cluster aggregation failures.
    \item We conduct a blinded process audit showing that numerical correctness and critic-labeled derivational support can diverge sharply, while explicitly limiting this conclusion to the constructed audit pool.
\end{itemize}

\section{Related work}

\paragraph{Formal geometry systems.}
Formal and neuro-symbolic geometry systems use explicit representations for deduction and checking.  FormalGeo supplies a formal language and symbolic environment for Olympiad geometry, and FGeo-HyperGNet adds learned theorem selection to formal deduction \cite{zhang2023formalgeo,zhang2024fgeo}.  AlphaGeometry combines a neural language model with symbolic deduction for synthetic geometry proofs \cite{trinh2024solving}.  Related work also studies autoformalization and reliable proof generation in Euclidean geometry \cite{murphy2024autoformalizing,sultan2025towards}.  Our setting isolates an earlier stage: an LLM receives an initial formal state and generates a textual derivation with a numerical answer.  We freeze these generated candidates and analyze them before symbolic search or checking can repair an error.

\paragraph{Test-time sampling and selection.}
Test-time sampling uses either repeated-answer aggregation or a learned score to select among multiple candidates.  Self-consistency uses answer frequency \cite{wang2022self}, while outcome verifiers rank sampled mathematical solutions \cite{cobbe2021training}.  Later studies examine repeated sampling and test-time compute at larger budgets \cite{brown2024large,snell2025scaling,montgomery2025budget}.  The shortfall between oracle candidate coverage and verifier-selected accuracy is commonly called the generation--verification gap \cite{saad2025shrinking}.  For verifier-based readouts, our $\Delta_R$ measures this shortfall on frozen answer clusters; we use the broader term generation-to-selection gap because $R$ also includes non-verifier readouts.

\paragraph{Verifier reliability and process supervision.}
Verifier behavior depends on its supervision target and on whether it scores one trace, a pair, or a set.  Process supervision can outperform outcome-only feedback in mathematical reasoning \cite{uesato2022solving,lightman2024let}, and verifier quality determines whether additional inference compute improves accuracy \cite{dorner2025roc,chen2026rethinking}.  Generative verification and cross-sample reasoning replace independent pointwise scores with richer comparisons \cite{shi2025heimdall,qi2025learning}, while verifier aggregation can reduce mismatch between generation and verification \cite{saad2025shrinking}.  We evaluate these alternatives on traces conditioned on the same formal geometry state.  The outcome verifier measures final-answer correctness and supplies scores for readout, whereas the blinded process critic evaluates derivational support on separately constructed representatives.  Keeping these roles separate prevents an answer-correctness score from being interpreted as proof certification.

\paragraph{Positioning of CRV.}
CRV's distinction lies in combining three controls in one evaluation protocol rather than claiming novelty for any component in isolation.  First, all readouts operate on identical frozen numerical-answer clusters, so differences among readouts are attributable to selection rather than generation.  Second, CRV conditions unrealized coverage on correct-answer multiplicity and within-problem hardest-negative discrimination, separating answer absence, candidate-ranking failures, and cluster-aggregation effects.  Third, it reports blinded, critic-labeled derivational support on constructed representatives as a separate axis rather than treating numerical correctness or outcome-verifier scores as proof certification.

\section{Approach}

\subsection{Coverage, realization, and validity evidence}

The CRV framework separately measures whether a sampled pool contains the correct answer (coverage), whether a readout selects it (realization), and what a blinded audit says about constructed representatives (validity evidence).  Its inputs are a dataset $D$ of problem-answer pairs, an LLM generator $G$, a per-problem sampling budget $K$, a deployable readout $R$, and a process auditor $V$.  For each problem, $G$ produces $K$ reasoning traces.  We cluster traces by numerical answer and freeze the resulting pool before $R$ selects one answer.  Separately from the deployed readout, the auditor $V$ evaluates constructed representatives from correct-answer and hard incorrect-answer clusters without access to their answer-correctness labels. We summarize the analysis with the CRV profile vector

\begin{equation}
 \mathcal{P}_{\mathrm{CRV}}=
 \bigl(\mathrm{pass@}K,\mathrm{Acc}(R),\Delta_R,\eta_R,
 \mathbf{b}_R,\rho_{\mathrm{audit}}\bigr),
\end{equation}
where $\mathbf{b}_R=[\mathrm{Acc}_R(c_1),\mathrm{Acc}_R(c_2),\mathrm{Acc}_R(c_{3+})]$ records multiplicity-band readout accuracy, and $\rho_{\mathrm{audit}}$ denotes the proportion of audited correct-answer representative traces labeled as supported by the critic. The $c_0$ band is omitted from $\mathbf{b}_R$ because its readout accuracy is zero by construction. The profile vector keeps coverage, realization, and validity evidence as separate measurements.

\subsection{Frozen candidate-pool construction}

Construction of a frozen candidate pool gives every readout the same answer-level decision space. Each problem $x$ contains a formal geometry state, a query, and a gold numerical answer $y$.  The LLM generator samples $K$ textual traces $r_1,\ldots,r_K$, and a deterministic parser extracts answers $a_1,\ldots,a_K$ from all traces.  We group answers into clusters $C_1,\ldots,C_m$ based on their numerical value while preserving which traces belong to each answer cluster.  The candidate pool is frozen before being passed to the readouts, so any difference comes from the readout rather than from different generations. The answer-level coverage is defined as
\begin{equation}
 \mathrm{pass@}K=\frac{1}{N}\sum_{n=1}^{N}
 \mathbb{1}\!\left[\exists i\leq K:a_{n,i}=y_n\right].
\end{equation}
Coverage only indicates whether the sampled candidate pool contains the gold numerical answer.  It is an upper bound for a readout on that pool, not evidence that the pool contains a valid mathematical proof.

\subsection{Coverage realization}

A readout $R$ maps a frozen candidate pool to one predicted answer
cluster. We evaluate one training-free readout and several verifier-based readouts.
The training-free Majority readout selects the largest answer cluster. Verifier-based readouts first score the candidates and then aggregate
those scores at the answer-cluster level. Specifically, a pointwise verifier assigns each candidate a probability $q_i$. Best-of-$K$ returns the answer associated with the highest-scoring candidate. We then obtain the calibrated probability
$
p_i=\sigma\!\left(\operatorname{logit}(q_i)/\tau\right)
$
through temperature scaling.
The WSC readout uses $p_i^\alpha$ as the candidate weight and selects
\begin{equation}
  \hat{C}_{\mathrm{WSC}}
  = \arg\max_{C_j}\sum_{i:r_i\in C_j} p_i^{\alpha},
\end{equation}
where $\alpha$ is tuned on the calibration set while keeping the sampled candidates and answer clusters fixed. Other readouts, including Best-of-$K$, pairwise, generative, and
set-level selectors, also operate on the same frozen candidates.

For readout $R$, realized accuracy is $\mathrm{Acc}(R)$, and the \emph{generation-to-selection gap} is defined as
\begin{equation}
 \Delta_R(K)=\mathrm{pass@}K-\mathrm{Acc}(R).
\end{equation}
We define the realization ratio as the fraction of the potential improvement between majority voting and $\mathrm{pass@}K$ achieved by the readout:
\begin{equation}
\eta_R=\frac{\mathrm{Acc}(R)-\mathrm{Acc}(\mathrm{Maj})}
 {\mathrm{pass@}K-\mathrm{Acc}(\mathrm{Maj})}.
\end{equation}
We report $\eta_R$ only when $\mathrm{pass@}K>\mathrm{Acc}(\mathrm{Maj})$. These metrics separate generation failures from selection failures on covered problems.

\subsection{Multiplicity-conditioned failure localization}
Correct-answer multiplicity provides a first step toward localizing the failures hidden by aggregate accuracy. For problem $n$, we define the correct-answer multiplicity as
$s_n=\sum_i\mathbb{1}[a_{n,i}=y_n]$.
Based on this multiplicity, we divide problems into four bands, denoted by $c_0$, $c_1$, $c_2$, and $c_{3+}$, corresponding to zero, one, two, or at least three correct-answer candidates. These bands separate different failure regimes. A $c_0$ problem is a generation failure because the sampled pool contains no correct answer. By contrast, $c_1$ and $c_2$ are low-multiplicity cases, whereas $c_{3+}$ provides repeated answer-level evidence for the correct answer.

Although this decomposition shows how much correct-answer evidence is available, it does not reveal whether the verifier can distinguish that evidence from the strongest competing error. In particular, even when a problem is covered, the verifier may assign a higher score to an incorrect trace than to every correct trace. To measure this within-problem discrimination directly, we define the hardest-negative margin for each covered problem $n$ as
\begin{equation}
 m_n=\max_{i:a_{n,i}=y_n}q_{n,i}
      -\max_{i:a_{n,i}\neq y_n}q_{n,i}.
\end{equation}

Because temperature scaling with $\tau>0$ is strictly increasing, $q_i$ and $p_i$ induce identical within-problem rankings and hardest-negative win indicators; margin magnitudes are reported on the uncalibrated $q_i$ scale. A positive margin indicates that at least one correct trace is scored above all incorrect traces. We compute the margin only for covered pools containing at least one scored incorrect candidate; all-correct pools are readout successes but have no hardest negative. Aggregating this comparison across eligible problems gives the win rate $\mathbb{P}(m_n>0)$, which measures how often the verifier ranks a correct trace above the strongest competing error within the same problem. This metric isolates candidate-level discrimination and directly characterizes the decision made by Best-of-$K$. For cluster-aggregating readouts such as WSC, realized accuracy additionally depends on how candidate scores are combined within each answer cluster. Unlike global candidate-level AUC (area under the receiver operating characteristic curve), which pools candidate comparisons across different problems, the hardest-negative win rate provides a problem-level diagnostic of whether the verifier can identify a correct trace among its strongest competitors.

Interpreting this win rate requires an appropriate random-ranking reference. The baseline cannot be fixed at $0.5$, because the probability that a correct candidate ranks first depends on how many correct candidates are present in the pool. Under exchangeable candidate scores, every scored candidate is equally likely to receive the highest score. Therefore, the probability that the top-ranked candidate is correct is equal to the proportion of scored candidates that are correct. The band-level multiplicity-adjusted null averages this fraction per problem. We compare the observed hardest-negative win rate with this null rather than with a universal reference value of $0.5$.

\subsection{Blinded derivational-support audit}

A correct answer does not guarantee a valid derivation. We therefore audit whether constructed representative traces are supported by the given formal state, independently of their final-answer correctness. Rather than auditing the deployed output of a particular readout, CRV selects one representative from a correct-answer cluster and one hard representative from a competing incorrect-answer cluster for each audited covered problem. The auditor evaluates each trace separately without access to the gold answer, correctness label, competing trace, or verifier score. It returns a verdict of \emph{supported}, \emph{refuted}, or \emph{uncertain}, together with a confidence label of \emph{high} or \emph{low} and, for a refuted trace, the first identified invalid step. We score supported/high, supported/low, uncertain, refuted/low, and refuted/high as $2$, $1$, $0$, $-1$, and $-2$, respectively; a higher, equal, or lower correct-representative score yields a win, tie, or loss. A stratified secondary audit checks selected verdicts using a second model or human review. Because only selected correct-answer representatives are used to compute $\rho_{\mathrm{audit}}$, it is a critic-conditional supported rate for this subset rather than an estimate of derivational validity across all generated traces.

\section{Experiments}

We evaluate CRV on legacy and shifted formal-geometry states. We measure coverage, realization, and validity evidence, including performance patterns conditioned on correct-answer multiplicity and critic-labeled derivational support for constructed representatives.

\subsection{Experimental setup}

\paragraph{Datasets and split discipline.}
Table~\ref{tab:training-data} summarizes the datasets used to
train the generator and verifier.
\begin{table}[!h]
\centering
\small
\setlength{\tabcolsep}{3.1pt}
\begin{tabular}{@{}lrrr@{}}
\toprule
Training set & $N$ & E/M/H & Examples \\
\midrule
Generator training & 2,556 & 479/847/1,230 & 4,901 \\
Verifier training & 3,338 & 607/1,049/1,682 & 31,701 \\
\bottomrule
\end{tabular}
\caption{Training-set composition}
\label{tab:training-data}
\end{table}
In Table~\ref{tab:training-data}, $N$ is the number of unique
problems.  The Examples column counts correct-answer
traces for generator SFT and labeled candidates for verifier
training.  E/M/H reports the numbers of easy, medium, and hard
problems, respectively. The generator and verifier are trained on task-specific corpora with substantial problem-level overlap: the generator uses only filtered correct-answer traces, whereas the verifier requires both correct and incorrect candidates for outcome discrimination. Table~\ref{tab:datasets} summarizes the evaluation datasets.
\begin{table}[!h]
\centering
\small
\setlength{\tabcolsep}{3.1pt}
\begin{tabular}{@{}lrrrrr@{}}
\toprule
Evaluation set & $N$ & E/M/H & Unsolved (\%) & Steps & Nodes \\
\midrule
Legacy E/M & 386 & 127/259/0 & 0.0 & 1.97 & 12.16 \\
Legacy hard & 184 & 0/0/184 & 0.0 & 5.34 & 14.19 \\
HardShift441 & 441 & 0/0/441 & 92.1 & 5.84 & 18.70 \\
\bottomrule
\end{tabular}
\caption{Evaluation-set composition}
\label{tab:datasets}
\end{table}

In Table~\ref{tab:datasets}, the Unsolved column denotes the percentage of problems
unsolved by the reference solver, while Steps and Nodes report the
mean dataset-annotated proof depth and the mean number of nodes in the initial formal
state. HardShift441 denotes a solver-unsolved hard-state shift rather than a claim that every problem was globally unseen. All 441 states retain numerical targets inherited from FormalGeo7K, whose source paper reports manual formalization, annotated solution theorem sequences, and solver validation of all annotated problems \cite{zhang2023formalgeo}. To construct the state records, we use a reference HyperGNet run that solves 35 problems and leaves 406 unsolved; by contrast, it solves all 184 legacy-hard problems. This run's failure is therefore used as a shift indicator, not as evidence of a missing or unvalidated dataset label. The two sets have no overlapping problem identifiers, and a HardShift441 state has 4.51 more nodes and about 0.5 more proof steps on average.

HardShift441 was frozen before generation of its primary candidate pools. Subsequent exploratory experiments split it by problem identifier into 300 adaptation, 60 calibration, and 81 development problems. The primary full-set pools predate this split; later split-specific results are labeled exploratory. Within each reported split-specific experiment, manifests enforce separation from its designated training data; historical reuse prevents a stronger globally unseen claim.

\paragraph{Generators, verifiers, and readouts.}

We use a LoRA-adapted Qwen2.5-7B model \cite{qwen2025qwen25} as the main generator $G$, while the scaling comparison uses Qwen2.5-14B.  The cross-backbone development analysis evaluates all combinations of Qwen2.5-7B, Qwen3-8B \cite{yang2025qwen3}, and Llama3-8B \cite{grattafiori2024llama} as generator and verifier backbones, using three training seeds per verifier; we additionally evaluate a Qwen3-32B verifier. Majority is training-free, whereas Best-of-$K$ and WSC use the trained verifiers.

\paragraph{Process-audit instantiation.}

The process audit is conducted on the 195 covered problems in the 300-problem HardShift441 adaptation partition. The process critic is GPT-5.4; a stratified secondary check uses Claude Fable 5 or human review. For each problem, we construct a pair of representative traces: one from a correct-answer cluster and one from a competing incorrect-answer cluster. The correct-answer representative is the highest-scoring complete trace in the most populous answer cluster whose value matches the gold answer within the stated tolerance, while the incorrect-answer representative is selected from the highest-scoring incorrect cluster. These pairs are constructed specifically for process auditing and are independent of the traces selected by the deployed readout.

\paragraph{Candidate generation and parsing.}

The generator $G$ samples 16 candidates with temperature 0.8, top-$p$ 0.95, and a 3,072-token limit unless noted otherwise.  A deterministic parser extracts structured or boxed numbers and compares them at absolute and relative tolerances of $10^{-6}$.  The budget curves use observed prefixes $K\in\{1,2,4,8,16\}$.

\paragraph{Verifier training and statistical analysis.}

The main verifier is a LoRA-adapted Qwen2.5-7B model with rank 16, scaling factor 32, and dropout 0.05.  Each example consists of a problem representation and a single trace. It is labeled as \emph{Yes} if the answer extracted from the trace matches the correct answer, and \emph{No} otherwise. We train the verifier for one epoch on 31,701 labeled state--trace pairs, with a learning rate of $10^{-4}$, a maximum sequence length of 4,096, and an effective batch
size of 16. For each pair, the verifier score $q_i$ is defined
as the normalized first-token probability assigned to
\emph{Yes}. Calibration is
performed separately for three verifier seeds, yielding
$(\tau,\alpha)$ values of $(1.5,2)$, $(1.5,2)$, and
$(1.75,1)$. Reported HardShift441 results are averaged over the nine
matched combinations of three generator seeds and three
verifier seeds.
For the Qwen3-8B replication, we use three generation seeds on
the legacy splits. Results at smaller sampling budgets are
computed from prefixes of the same 16-candidate pools, while
the verifier and its calibration parameters remain fixed.
Unless otherwise noted, primary 95\% confidence intervals
are computed using 10,000 percentile-bootstrap resamples at
the problem-identifier level. Pairwise binary comparisons on
the same problems use two-sided exact McNemar tests. The primary
bootstrap intervals condition on the trained generator and verifier
seeds. In a separate verifier-scaling
experiment, the reported Qwen3-32B verifier result is averaged over
two training seeds.

\subsection{Repeated sampling reveals answer-level coverage}

\begin{table*}[!t]
\centering
\small
\setlength{\tabcolsep}{5.2pt}
\begin{tabular}{@{}lrrrrrrc@{}}
\toprule
Setting & pass@16 & Majority & Best-of-$K$ & WSC & $\Delta_{\mathrm{WSC}}$ & $\eta_{\mathrm{WSC}}$ & $\mathbf{b}_{\mathrm{WSC}}$ \\
\midrule
Legacy hard & 79.35 & 42.93 & 52.17 & 53.44 & 25.91 & 28.86 & 7.69/38.89/87.58 \\
HardShift441 & 68.93 & 33.56 & 36.51 & 37.99 & 30.94 & 12.54 & 4.87/13.53/75.63 \\
\bottomrule
\end{tabular}
\caption{Coverage and realization components of CRV at $K=16$. Rates are percentages; $\mathbf{b}_{\mathrm{WSC}}$ lists accuracy on $c_1/c_2/c_{3+}$.}
\label{tab:crv-profiles}
\end{table*}

\begin{figure}[t]
  \centering
  \includegraphics[width=\columnwidth]{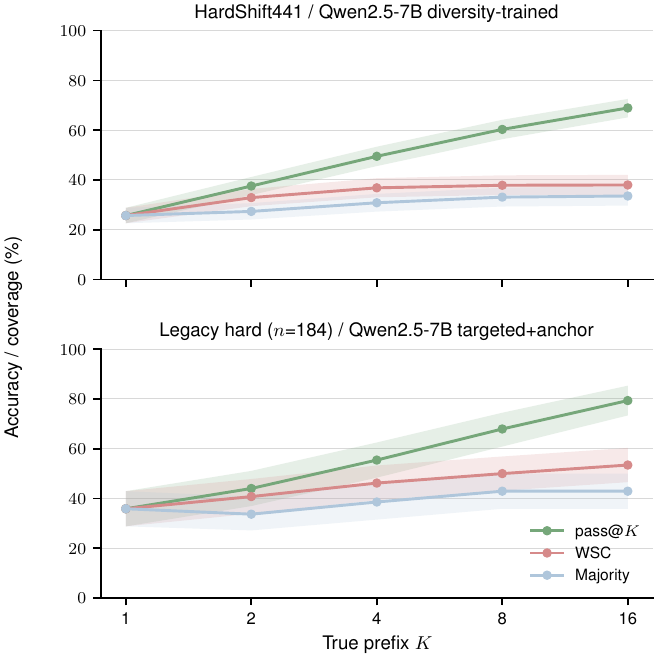}
  \caption{True-prefix budget curves with fixed calibration and pointwise 95\% problem-identifier bootstrap intervals.  The HardShift441 panel uses the Qwen2.5-7B diversity-trained generator, whereas the legacy-hard panel uses an earlier Qwen2.5-7B targeted-distillation generator trained with the same number of anchor examples that the base model answered correctly.  Because the generators differ, the panels show within-setting trends rather than a direct domain comparison.  In both settings, coverage continues to rise after WSC nearly plateaus.}
  \label{fig:kcurve}
\end{figure}

Repeated sampling finds answers that a single sampled output misses, but the readouts recover only part of this coverage. At $K=16$, average single-sample accuracy---the mean candidate correctness over 16 candidates per problem, 441 problems, and three generator seeds---is 24.17\% on HardShift441. Table~\ref{tab:crv-profiles} summarizes the $K=16$ profiles, and Figure~\ref{fig:kcurve} shows the budget curves. WSC realizes 28.86\% of the majority-to-coverage headroom on legacy hard but only 12.54\% on HardShift441. Increasing $K$ from 8 to 16 raises HardShift441 pass@$K$ from 60.32\% to 68.93\%, while WSC remains nearly unchanged at 37.87\% and 37.99\%. Consequently, $\Delta_{\mathrm{WSC}}(K)$ grows by 8.49 percentage points (95\% bootstrap confidence interval (CI) $[6.60,10.43]$). Using a Qwen2.5-14B generator raises pass@16 by 2.80 percentage points but WSC accuracy by only 0.58 percentage points (paired 95\% CI $[-1.36,2.54]$). On legacy easy/medium problems, the base Qwen2.5-7B model obtains 51.81\% greedy accuracy and 89.90\% pass@16. Among the 186 problems missed by greedy decoding, the sampled pools recover the correct answer for 79.03\%.

An exploratory cross-backbone analysis on the reused development split yields the same pattern.  Candidate pools from Qwen2.5-7B, Qwen3-8B, and Llama3-8B all show substantial gaps between pass@16 and both majority voting and the learned readout.  Switching the LLM generator changes the magnitude of $\Delta_R$ but does not eliminate it on this split.

\subsection{Selectors leave much of the coverage unrealized}

\paragraph{Unrealized coverage concentrates at low correct-answer multiplicity.}

\begin{figure}[t]
  \centering
  \includegraphics[width=\columnwidth]{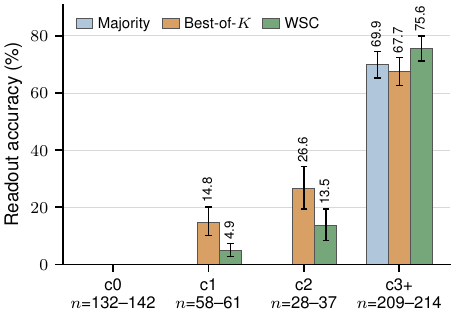}
  \caption{HardShift441 readout accuracy by correct-answer multiplicity among 16 frozen samples.  Values average nine generator-seed $\times$ verifier-seed cells; error bars are pointwise 95\% problem-identifier bootstrap intervals and labels give the per-cell count range. Absent bars denote zero accuracy; $c_0$ is zero by construction because the correct answer is absent.}
  \label{fig:support}
\end{figure}

The multiplicity bands in Figure~\ref{fig:support} help distinguish candidate-level discrimination from cluster-level score aggregation. Best-of-$K$ achieves accuracies of 14.76\% on $c_1$ and 26.59\% on $c_2$, whereas WSC reaches only 4.87\% and 13.53\%, respectively. Relative to Best-of-$K$, WSC performs worse by 9.89 percentage points on $c_1$ (95\% CI $[-14.01,-6.27]$) and by 13.06 percentage points on $c_2$ (95\% CI $[-18.96,-7.63]$). This ordering reverses in the $c_{3+}$ band, where WSC achieves 75.63\% accuracy and outperforms Best-of-$K$ by 7.97 percentage points (95\% CI $[4.99,11.08]$). These results are consistent with two difficulties in low-multiplicity pools: pointwise scoring often fails to identify a rare correct trace, and frequency-weighted aggregation further downweights correct-answer clusters with multiplicity one or two. At higher multiplicity, the same aggregation strategy is instead associated with higher accuracy. All readouts have zero accuracy on $c_0$ by construction because the candidate pool contains no correct answer.

The within-problem comparison reduces, but does not eliminate, confounding by static problem identity. Among the 75 problems that fall into different multiplicity bands across random seeds, WSC accuracy is 24.67 percentage points higher when the same problem appears in $c_{3+}$ rather than in $c_1$ or $c_2$ (95\% CI $[16.44,32.89]$). The hardest-negative margin is also higher by 0.091 (95\% CI $[0.045,0.136]$), showing an association between higher correct-answer multiplicity and stronger verifier separation from the hardest incorrect trace. This comparison does not estimate a causal effect of multiplicity because changing the random seed affects both the number and the quality of the generated traces. The same descriptive pattern appears in all 27 cells of the cross-backbone development analysis. Restricting the summary to each verifier's matching generator pool and averaging over its three training seeds, WSC accuracy ranges from 2.78\% to 7.41\% on $c_1$ and 25.92\% to 33.33\% on $c_2$, compared with 80.70\% to 90.20\% on $c_{3+}$.

\paragraph{Within-problem discrimination is lower on the shifted set.}

Candidate-level AUC is lower on HardShift441 than on the in-domain evaluation data, decreasing from approximately 0.903 to 0.819. Because the datasets and candidate-generating systems differ, this comparison is descriptive and does not identify the shift as the cause. The hardest-negative win rates for $c_1$, $c_2$, and $c_{3+}$ are 14.57\%, 26.59\%, and 66.82\%, respectively, compared with multiplicity-adjusted random-ranking baselines of 6.25\%, 12.50\%, and 45.71\%. The observed win rates exceed these baselines by 8.32 percentage points for $c_1$ (95\% CI $[3.58,13.56]$) and 14.09 percentage points for $c_2$ (95\% CI $[6.83,21.66]$). Thus, although the verifier performs better than random ranking, it still fails to rank a correct trace above the strongest incorrect trace in most low-multiplicity cases. Aggregate candidate-level AUC and the pooled win rate of 52.2\% hide this weakness.
This discrimination pattern coincides with limited gains from larger candidate pools. Larger candidate pools can introduce both previously missing correct answers and stronger incorrect competitors. When the sampling budget increases from 8 to 16 candidates, pass@$K$ grows substantially faster than realized accuracy, which is consistent with the readout making limited use of the additional answer coverage in this setting. Therefore, for a fixed candidate pool, coverage should be interpreted as an upper bound on achievable readout accuracy rather than as an estimate of deployed performance.

\subsection{A blinded critic labels most audited correct-answer representatives as refuted}

\begin{figure}[t]
  \centering
  \includegraphics[width=\columnwidth]{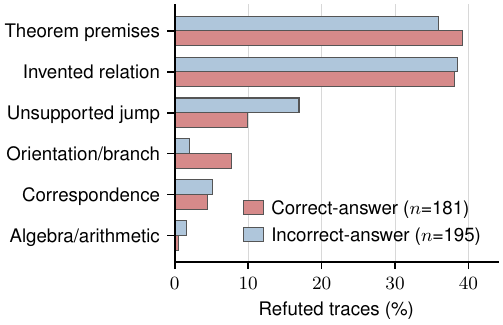}
  \caption{First identified errors among the 181 correct-answer and 195 incorrect-answer representatives labeled as refuted by the blinded process critic.  Percentages are normalized separately for each group.  Because the paired representatives were selected for audit, these distributions are conditional critic outputs rather than estimates of valid-trace prevalence.}
  \label{fig:process}
\end{figure}

Figure~\ref{fig:process} shows the process critic's highly skewed verdict distribution. Among 195 correct-answer representatives, it labels 12 supported (6.15\%, Wilson 95\% CI $[3.55,10.45]\%$), 181 refuted---179 with high confidence and two with low confidence---and two uncertain. It labels none of the 195 hard incorrect-answer representatives as supported; all are labeled as refuted, so 376 of 390 audited traces (96.4\%) receive a refuted label. This makes $\rho_{\mathrm{audit}}$ a critic-conditional supported rate for the constructed correct-answer representatives, not an estimate of valid-trace prevalence. Under the confidence-aware ordinal score, the correct-answer representative wins 16 pairs, loses two, and ties 177; counting each tie as 0.5 yields a tie-adjusted win rate of 53.6\%.

Targeted agreement checks detect little disagreement in the reviewed strata: human review agrees with the critic on all 20 reviewed correct-answer verdicts (12 supported and eight high-confidence refutations), and a second model agrees with the critic on 33 of 34 reviewed verdicts. These checks do not establish critic calibration, full-pool accuracy, valid-trace prevalence, or localization accuracy. Approximately 12\% of refutations involve configuration facts that the textual state does not fully determine, a potential source of over-refutation; common remaining errors include invented relations and theorem applications with unmet premises. In this constructed pool, the process audit provides little answer-selection signal; its possible use for abstention or certification remains a direction for future evaluation.

\section{Exploratory intervention screening}

\begin{table*}[!t]
\centering
\small
\setlength{\tabcolsep}{3pt}
\begin{tabular}{@{}
  >{\centering\arraybackslash}p{0.10\textwidth}
  >{\raggedright\arraybackslash}p{0.16\textwidth}
  >{\raggedright\arraybackslash}p{0.19\textwidth}
  >{\raggedright\arraybackslash}p{0.16\textwidth}
  >{\raggedright\arraybackslash}p{0.33\textwidth}
@{}}
\toprule
\textbf{Group}
& \textbf{Intervention}
& \textbf{Dataset / split}
& \textbf{Metric(s)}
& \textbf{Main result} \\
\midrule

\multirow[c]{3}{=}[-25pt]{%
  \centering\arraybackslash\textbf{Generation}%
}
& \textbf{Generator--budget trade-off}
& Legacy hard + E/M (reused)
& WSC accuracy
& Qwen3-8B $K=8$ vs.\ Qwen2.5-7B diversity-trained $K=16$:
$+7.07$ pp (hard); $+2.94$ pp (E/M) \\
\addlinespace[2pt]

& \textbf{Increased sampling}
& Legacy hard + E/M (reused)
& WSC accuracy
& Qwen3-8B $K=16$ vs.\ $K=8$:
$+1.93$ pp (hard); $+1.76$ pp (E/M) \\
\addlinespace[2pt]

& \textbf{Generator mixture}
& HardShift441 development (81 problems, reused)
& pass@16; best evaluated readout accuracy
& 12 Qwen3-8B + 4 Qwen2.5-7B vs.\ 16 Qwen3-8B:
pass@16 $80.25\%\!\to\!81.01\%$;
accuracy $59.26\%\!\to\!57.88\%$ \\

\midrule

\multirow{5}{=}[-25pt]{\centering\textbf{Selection}}
& \textbf{Trained pointwise verifier}
& Legacy hard
& WSC accuracy
& Majority 42.93\%; verifier-based WSC 53.44\%
($+10.51$ pp) \\
\addlinespace[2pt]

& \textbf{Qwen3-32B verifier}
& Legacy hard
& WSC accuracy
& Qwen2.5-7B verifier 53.44\%;
Qwen3-32B verifier 57.61\% ($+4.17$ pp) \\
\addlinespace[2pt]

& \textbf{Groupwise rank training}
& HardShift441 development (81 problems) + legacy hard
& WSC accuracy
& HardShift441 development: $+4.93$ pp; legacy hard:
$-9.24$ pp \\
\addlinespace[2pt]

& \textbf{Set selector with traces}
& E/M external development (9,200 $K=4$ subsets)
& Selection accuracy
& WSC 82.98\%; set selector 82.27\%
($-0.71$ pp) \\
\addlinespace[2pt]

& \textbf{Generative answer critic}
& HardShift441 development/calibration (125 answer pairs)
& Correct-answer preference
& Correct answer receives the better verdict in 85 pairs;
26 ties \\

\midrule

\multirow{1}{=}{\centering\textbf{Aggregation}}
& \textbf{Verifier ensemble}
& HardShift441 development (81 problems, reused)
& $K=8$ WSC accuracy
& Three-seed ensembles improve by $+1.15$ to $+2.58$ pp
over the mean of the three individual verifiers \\

\bottomrule
\end{tabular}
\caption{Exploratory intervention-screening results. Rates are percentages and
differences are percentage points (pp). Each row reports an independent
comparison on the indicated dataset and split; results should not be compared
across rows.}
\label{tab:interventions}
\end{table*}

Table~\ref{tab:interventions} screens changes to generation, selection, and aggregation. Its rows use different candidate pools and splits, so the results are exploratory rather than directly comparable.

\paragraph{Several interventions yield local gains.}
Stronger generation improves WSC on both reused legacy splits, and the trained pointwise verifier improves legacy-hard accuracy over Majority by 10.51 percentage points. Larger verifiers and verifier ensembles also yield local gains. These results apply only to the reported candidate pools and splits.

\paragraph{No tested change closes the gap consistently.}
Additional sampling realizes only part of its added coverage, while the generator mixture raises coverage but lowers readout accuracy. Groupwise ranking improves HardShift441 development but regresses on legacy hard, and the set selector trails WSC. No tested intervention therefore closes the gap consistently across settings.

\section{Discussion}

\paragraph{Interpretation and implications.}
On HardShift441, unrealized answer-level coverage concentrates in low-multiplicity pools: when the correct answer appears once or twice, the evaluated readouts rarely select it. The verifier may favor a convincing incorrect trace, while WSC also favors frequent answers. This pattern is consistent with rare correct answers being outweighed, but does not establish a causal multiplicity effect.
The results motivate stronger within-problem discrimination and aggregation that does not suppress low-multiplicity clusters. Because the critic refutes most audited correct-answer representatives, answer selection remains distinct from derivational support and requires formal checking or geometry-specific process models.

\paragraph{Limitations.}
Our conclusions concern numerical-answer generation over formal states, not symbolic proof search. The multiplicity analysis is descriptive, and the process audit covers selected rather than deployed traces. Although the targets inherit FormalGeo7K's source validation, we do not independently audit them; the reference HyperGNet run leaves 406 of 441 problems unsolved. The bootstrap intervals condition on the trained generator and verifier seeds and exclude retraining uncertainty.

\paragraph{Use of AI.}
LLMs generate solutions, score candidates, and audit derivations; their outputs are model signals, not ground truth or proof certificates. Generative AI provided limited editorial assistance; the authors verified the analysis and retain responsibility for the manuscript.

\section{Conclusion}

CRV separates coverage, realization, and validity evidence on frozen candidate pools. On HardShift441, pass@16 reaches 68.9\% while WSC reaches 38.0\%, with particularly low readout accuracies among covered problems in $c_1$ and $c_2$. Thus, pass@$K$ should not be interpreted as either deployable readout accuracy or evidence of derivational validity; the results motivate further work on low-multiplicity selection and geometry-specific verification.

\bibliography{formalgeo_generation_selection}

\clearpage
\includepdf[pages=-,fitpaper=true]{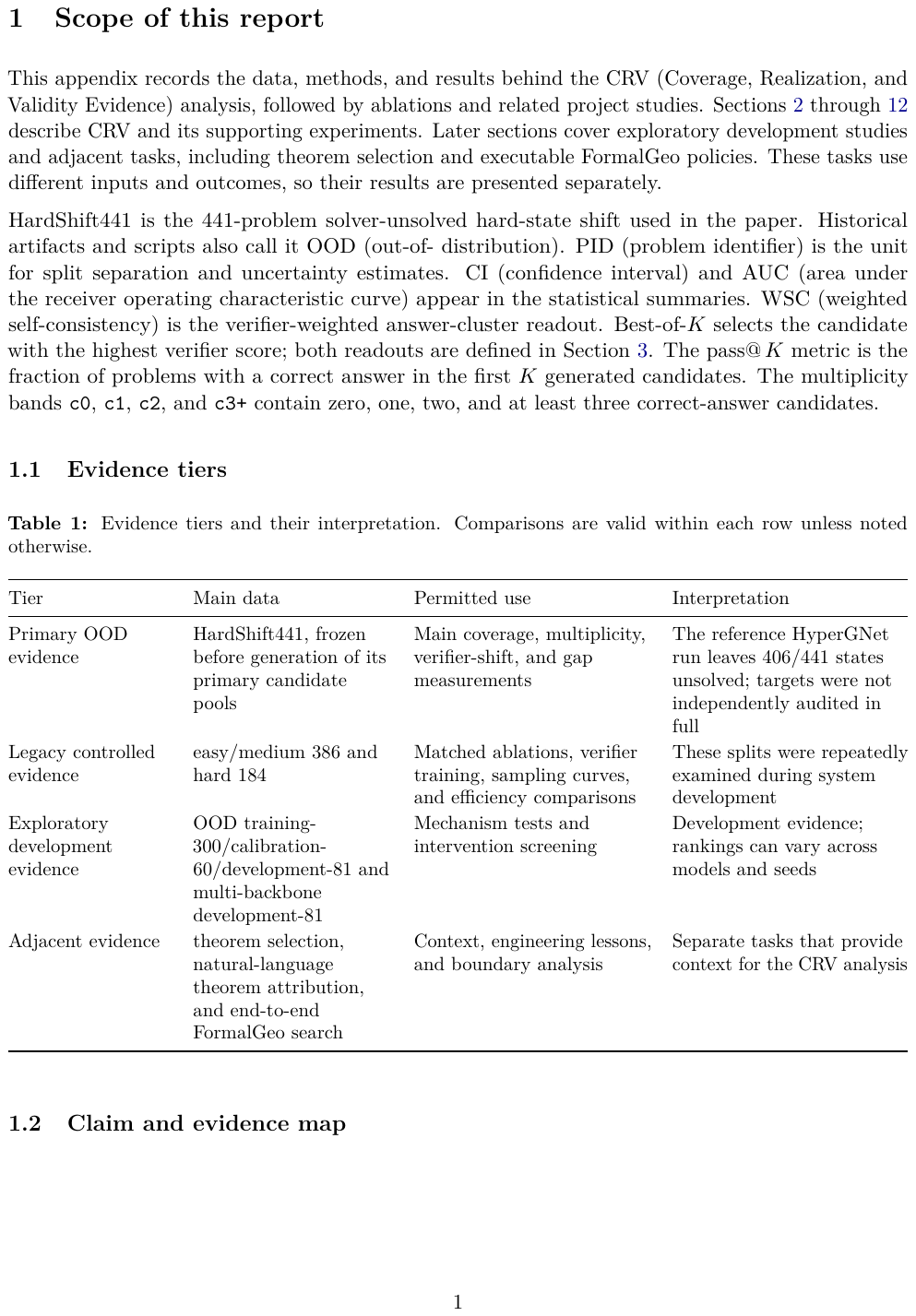}

\end{document}